\pdfoutput=1

\documentclass[11pt]{article}

\usepackage[]{acl}

\usepackage{times}
\usepackage{latexsym}

\usepackage[T1]{fontenc}

\usepackage[utf8]{inputenc}

\usepackage{microtype}

\usepackage{graphicx}
\usepackage{multirow}
\title{TERMinator: A system for scientific texts processing}

\author{Elena Bruches \\
  \small{A.P. Ershov Institute of Informatics Systems / Russia} \\
  \small{Novosibirsk State University / Russia} \\
  \small{\texttt{bruches@bk.ru}} \\\And
  Olga Tikhobaeva \\
  \small{Novosibirsk State University / Russia} \\
  \small{\texttt{o.tikhobaeva@g.nsu.ru}} \\
  \AND
  Yana Dementyeva \\
  \small{Novosibirsk State University / Russia} \\
  \small{\texttt{y.dementeva@g.nsu.ru}} \\\And
  Tatiana Batura \\
  \small{A.P. Ershov Institute of Informatics Systems / Russia} \\
  \small{\texttt{tbatura@iis.nsk.su}}
  }

\begin{document}
\maketitle

\begin{abstract}
This paper is devoted to the extraction of entities and semantic relations between them from scientific texts, where we consider scientific terms as entities. In this paper, we present a dataset that includes annotations for two tasks and develop a system called TERMinator for the study of the influence of language models on term recognition and comparison of different approaches for relation extraction. Experiments show that language models pre-trained on the target language are not always show the best performance. Also adding some heuristic approaches may improve the overall quality of the particular task. The developed tool and the annotated corpus are publicly available at \href{https://github.com/iis-research-team/terminator}{https://github.com/iis-research-team/terminator} and may be useful for other researchers.

\end{abstract}

\section{Introduction}  

Nowadays the amount of scientific publications is constantly growing. In this regard, the processing of scientific texts becomes especially relevant in relation to rapidly developing scientific fields, for example, computer science. Information extraction from scientific texts can be useful in domain-specific areas, for completion of knowledge graphs, in search and question-answering systems. This paper describes the study on entity recognition and relation extraction from scientific texts on computer science in Russian.

Currently, there are a number of datasets with annotations of entities and relations in a general domain \cite{doddington-etal-2004-automatic,roth-yih-2004-linear,loukachevitch2021nerel}, biomedical domain \cite{kim2003genia,10.1016/j.jbi.2012.04.008,10.1093/database/baw068}, or even multi-domains \cite{RigoutsTerryn2020InNU}. Still it is more difficult to find a publicly available dataset such as SciERC \cite{luan-etal-2018-multi} for scientific fields other than biomedical, and especially in languages other than English.

Despite that the named entity recognition task is well studied, it still faces multiple challenges \cite{Li2022ASO}, namely, NER in domain-specific areas \cite{weber2021hunflair}, NER from noisy data \cite{derczynski2017results}, code-mixed data \cite{fetahu2021gazetteer}, and detection of fine-grained and nested named entities \cite{kim2021fine,ringland2019nne,loukachevitch2021nerel}. This is caused by several issues: defining boundaries of compound terms; recognition of whether a lexical unit is part of a compound term; identification of a lexical unit as a term depending on the context and topic of the text in which this lexical unit is used etc. The relation extraction task also remains an unsolved problem, as it often requires the use of knowledge outside the text (for example, from knowledge bases or obtained in another way), and also due to the lack of a large amount of labeled data in different languages.

Selection of the most appropriate language model, which is able to provide the best quality for extraction of terms and relations from scientific texts is one of the relevant issues. Our experimental results not only show the usefulness of the proposed dataset, but also provide baselines for further research.

We make the following contributions:
\begin{itemize}
    \item 
    Provide a new dataset for both tasks (term recognition and relation extraction) for Russian scientific texts and develop a TERMinator tool for further research experiments.
    \item
    Study of influence of language models (without additional information, with heuristics and dictionaries) on term extraction. 
    \item
    Compare three approaches for relation extraction (based on lexical patterns, classification with a CLS-vector, and combination of them).
\end{itemize}



\section{Related Work}

Entity recognition and relation extraction are the main tasks in information extraction. There are various approaches to solve them. 

A traditional approach includes two stages: extracting n-grams which potentially may be terms, and then classification whether this n-gram is a term or not. In \cite{stankovic-etal-2016-rule} authors proposed to use dictionaries and morphological and syntax information. There are some works which use pre-defined ontologies for terms extraction \cite{ivanisenko-etal-2020-anddigest}. Another idea is to solve this task as a sequence labeling \cite{kucza-etal-2018-term}. It allows to implement terms extraction in one stage and take into account syntax and semantic information from the context. For terms extraction the main challenge is to identify the exact term boundaries. In \cite{zhu-etal-2022-boundary} authors proposed to use boundary smoothing as a regularization technique to overcome this problem. 

Relation extraction is usually considered as a classification problem: for two given terms one needs to determine whether there is a semantic relation between them or not, and if they are related then to define its type. Some works describe the use of knowledge bases for relation extraction \cite{li-etal-2019-improving, baldini-soares-etal-2019-matching}. With the spread of transformers-based architectures, different pre-trained language models are used to solve this task \cite{shi-etal-2019-simple}. Some researchers try to make use of incorporating external data sources in the model, for example the list of hand-written syntax patterns \cite{tao-etal-2019-enhancing}, information about sentence syntax tree \cite{ningthoujam-etal-2019-relation, nayak-ng-2019-effective}.

Recently, special attention has been paid to systems which solve terms recognition and relation extraction jointly. The authors propose an architecture that sequentially extracts entities and relationships between them, but in end-to-end settings \cite{eberts-etal-2020-span, ji-etal-2020-span, huang-etal-2019-bert, miwa-bansal-2016-end}. Another idea is to train a model with two outputs: one output is for term extraction, and the other is for relation extraction \cite{xue-etal-2019-fine}. However such approaches require quite a lot of annotated data to find hidden regularities.

\section{Data Preparation}


For the experiments we create an annotated dataset, which consists of abstracts of scientific papers on Information Technology in Russian.

As entities we consider nouns or noun groups, which are terms in this particular domain. Terms that we recognize as entities may consist of one or several tokens (“\emph{software}”, “\emph{non-preemptive multitasking}”), abbreviations (“\emph{CPU}”, “\emph{DLL}”), names of programming languages (“\emph{Python}”, “\emph{C++}”) and libraries (“\emph{Pytorch}”, “\emph{SpaCy}”), hyphenated concepts containing Latin characters (”\emph{n-gram}”, ”\emph{web-service}”). Thereby we consider all possible chains of tokens that can be terms, except for those that are recursive or overlap. The entities are marked in the BIO format: each token is assigned a B-TERM tag if it is the initial tag for an entity, I-TERM if it is inside a term, or O if it is outside any entity.

Statistics for our dataset is presented in Table~\ref{tab:dataset}.

\begin{table}[ht]
\centering
\small
\begin{tabular}{lll}
\hline
\textbf{Unit} & \textbf{train} & \textbf{test}\\
\hline
\verb|texts| & {136} & {80}\\
\verb|tokens| & {12 809} & {11 157}\\
\verb|terms| & {2 028} & {2 027}\\ 
\verb|relations| & {356} & {620}\\ \hline
\end{tabular}
\caption{Dataset statistics}
\label{tab:dataset}
\end{table}

The list of relations is selected based on the following criteria. At first, a relation should be monosemantic (for example, we don’t consider a semantic relation \emph{<Entity-Destination>} because it has indirect meaning as well). Secondly, a relation should link scientific terms (for example, in relation \emph{<Communication-Topic>} (an act of communication is about topic) the actants are not scientific terms). Thus, six semantic relations were selected. Types of relations in a corpus, their meanings and distribution by train and test sets are presented in Table~\ref{tab:relationtypes}.

\begin{table}[hbt!]
\centering
\small
\begin{tabular}{llll}
\hline
\textbf{Relation type} & \textbf{Meaning} & \textbf{train} & \textbf{test}\\
\hline
\verb|CAUSE| & {x is the cause of y} & {19} & {19}\\ 
\verb|ISA| & {x is y} & {96} & {93}\\
\verb|PART_OF| & {x is part of y} & {23} & {87}\\ 
\verb|SYNONYMS| & {x is the same as y} & {35} & {22}\\ 
\verb|TOOL| & {x allows to create/etc. y} & {54} & {38}\\ 
\verb|USAGE| & {x is used for/in y} & {126} & {330}\\\hline
\end{tabular}
\caption{Types of relations}
\label{tab:relationtypes}
\end{table}

Here is a sample sentence where two terms and the relation between them are highlighted: \emph{Pokazany preimushchestva primeneniya <e1>mul'timedijnyh tekhnologij</e1> v <e2>uchebnom processe</e2> i effektivnost' ih ispol'zovaniya vo vremya lekcij i seminarov}. (The advantages of using multimedia technologies in the educational process and the effectiveness of their use during lectures and seminars are shown.).	The relation between \emph{e1} and \emph{e2} is \emph{USAGE}. 

The dataset is available for other researchers\footnote{\href{https://github.com/iis-research-team/ruserrc-dataset}{https://github.com/iis-research-team/ruserrc-dataset}}.




\section{Influence of language models on term extraction}
\subsection{Models without additional information}
\label{sec:without}

The experimental methodology is as follows: texts are fed as input; during the vectorization procedure, each text is divided into spans (in our case these are BPE tokens), each of which is assigned to a vector. Initially, the model learns to match tokens with labels by using the training data; then based on the revealed regularities the model makes predictions on the validation data. In this way, the metrics are fixed after each training epoch. The output is a set of labels associated by the system with each word from the input text.


We experimented with two models: multilingual BERT \cite{BERT2018} and BERT pre-trained on Russian texts \cite{ruBERT2019}.

In the first stage of the experiment each pre-trained language model was fine-tuned on the train set described above. The optimal learning rate was chosen as \(10^{-6}\), and the batch size was 12. Such values prevent overfitting and obtain the best results. At this stage, the metrics show that on partial match both models give the same performance, while in exact match the model pre-trained on Russian-language texts gives better scores.
 
Then we extended a train set by adding 212 texts with a pseudo labeling method. We collected a dictionary (list of scientific terms) in a semi-automatic way:
\begin{itemize}
    \item 
    We extracted 2-, 3- and 4-grams from the scientific papers and manually filtered phrases, which potentially can be terms. 
    \item
    We extracted all titles of articles from Wikipedia, which are included in a subgraph of category “Science”, and then manually selected words and phrases, which potentially can be terms.
\end{itemize}

Thus we obtained a list of 17 252 terms, which we used for pseudo labeling. This technique is useful for rapidly changing areas of knowledge, when it is difficult to have dictionaries of terms and keep them up to date. 
 
Due to the less detailed checking of the markup of the corpus, even with its comparatively large size, the metrics received with the models trained on it turned out to be lower than those of the same models trained on the manual annotated texts.

\subsection{Models with heuristics}
\label{sec:heur}
Experimentally we found out that in order to improve the quality of term extraction, we need to improve the definition of term boundaries.
The task of defining terms’ boundaries is more challenging than classifying a token as a “term” type. 
To improve the recognition of term boundaries, we apply some heuristics to handle such cases as removing a preposition as a first token of a term and some others. 
From the results shown in Table~\ref{tab:te-metrics}, it can be seen that the heuristics improve the quality of term extraction on the exact match.

\begin{table*}[hbt!]
\centering
\small
\begin{tabular}{llllllll}
\hline
\textbf{Train set} & \textbf{Model} & \textbf{F-M P} & \textbf{F-M R} & \textbf{F-M F1} & \textbf{P-M P} & \textbf{P-M R} & \textbf{P-M F1}\\
\hline
\multirow{6}{2cm}
{\texttt{Manually labeled}} & {\texttt{mBERT}} & {0.40} & {0.46} & {0.43} & \textbf{0.89} & \textbf{0.88} & \textbf{0.88}\\
& \verb|mBERT + h| & {0.49} & {0.45} & {0.47}  & {0.86} & {0.86} & {0.86}\\ 
& \verb|mBERT + d + h| & {0.47} & {0.50} & {0.48}  & {0.86} & {0.87} & {0.87}\\ 
& \verb|ruBERT| & {0.48} & {0.50} & {0.49} & \textbf{0.89} & \textbf{0.88} & \textbf{0.88}\\
& \verb|ruBERT + h| & \textbf{0.52} & {0.47} & {0.49}  & \textbf{0.89} & \textbf{0.88} & \textbf{0.88} \\ 
& \verb|ruBERT + d + h| & {0.49} & \textbf{0.51} & \textbf{0.50}  & {0.86} & {0.87} & {0.87} \\\hline 
\multirow{6}{2cm}
{\texttt{Pseudo labeled}} & {\texttt{mBERT}} & {0.33} & {0.34} & {0.34} & {0.80} & {0.75} & {0.75}\\
& \verb|mBERT + h| & {0.42} & {0.38} & {0.40}  & {0.80} & {0.79} & {0.79} \\ 
& \verb|mBERT + d + h| & {0.41} & {0.39} & {0.40}  & {0.80} & {0.80} & {0.80} \\ 
& \verb|ruBERT| & {0.32} & {0.32} & {0.33} & {0.78} & {0.76} & {0.75}\\
& \verb|ruBERT + h| & {0.40} & {0.37} & {0.38}  & {0.79} & {0.79} & {0.79} \\ 
& \verb|ruBERT + h + d| & {0.38} & {0.39} & {0.38}  & {0.79} & {0.74} & {0.76} \\\hline
\end{tabular}
\caption{Metrics for full match (F-M) and partial match (P-M) terms extraction; \emph{d} is for dictionary, \emph{h} is for heuristics.}
\label{tab:te-metrics}
\end{table*}

\subsection{Models with heuristics and dictionaries}

It the third stage, the system extracts terms not only with the trained model, but also with the use of the dictionary described above. Heuristics are also applied.


Table \ref{tab:te-metrics} shows that the ruBERT model fine-tuned on the manually annotated train set extracts the terms in half of the cases, which is the best result of a exact match. On the pseudo-labeled train set, this combined method gives good results for the multilingual BERT both for exact and partial matches.

Both models solve the task of term recognition with a high quality, which can be seen from the good results on a partial match. The best result on an exact match pertains to the RuBERT model supported by a dictionary and heuristics. The results of the models on the exact match are expectedly lower than the results on a partial match, which again draws our attention to the task of defining terms’ boundaries in texts.

The markup quality significantly affects the quality, as we can note from a comparison of the results of the first and second stages of the experiment (see Sections \ref{sec:without}, \ref{sec:heur}). Fine-tuning on the manually-annotated training set gives better performance than fine-tuning on the pseudo-labeled training set, even though its size is larger than the size of the manually-annotated one.
It is hard to compare results with other researchers as this is the first corpora for scientific texts in Russian as far as we know. But for the similar dataset in English SciERC \cite{luan-etal-2018-multi} the authors \cite{eberts-etal-2020-span} reported F1-measure to be 0.70.

We observed that for the term extraction task the model mistakes in recognizing the exact term boundaries. Another problem arises when a term is divided by other words or signs in the sentence, for example, \emph{"Morphological and syntax analysis"}. Probably, it should be solved at a post-processing stage.


\section{Comparison of approaches for relation extraction}
\subsection{Using lexical patterns}

At first, we applied an approach for relation identification based on lexical patterns. It consists in the following: for texts with tagged terms, we extract a context between each pair of terms, lemmatize it, and compare it with the lexical patterns. If they match, these two terms are connected by this relation. The length of the context should not exceed six words. This value was obtained experimentally by changing it and comparing the quality of the model. The obtained metrics for this approach are shown in Table~\ref{tab:metrics_lexicalpatterns}. We used 111 patterns; examples of patters and their distribution by relations are presented in Table~\ref{tab:lexicalpatterns}. 

\begin{table}[hbt!]
\centering
\small
\begin{tabular}{llll}
\hline
\textbf{Relation type} & \textbf{Precision} & \textbf{Recall} & \textbf{F1}\\
\hline
\verb|CAUSE| & {0.07} & {0.05} & {0.06}\\ 
\verb|ISA| & {0.18} & {0.19} & {0.19}\\
\verb|PART_OF| & {0.17} & {0.14} & {0.15}\\ 
\verb|SYNONYMS| & {0.23} & {0.82} & {0.35}\\ 
\verb|TOOL| & {0.06} & {0.08} & {0.07}\\ 
\verb|USAGE| & {0.21} & {0.39} & {0.27}\\
\verb|NO-RELATION| & {0.96} & {0.92}  & {0.94}\\ \hline
\verb|macro-average| & {0.27} & {0.37} & {0.29}\\ \hline
\end{tabular}
\caption{Metrics for lexical pattern’s approach}
\label{tab:metrics_lexicalpatterns}
\end{table}


\begin{table*}
\centering
\small
\begin{tabular}{p{0.09\linewidth} | p{0.38\linewidth} | p{0.38\linewidth} | p{0.04\linewidth}}
\hline
\textbf{Relation type} & \textbf{Examples of patterns (transliteration)} & \textbf{Examples of patterns (translation)} & \textbf{N}\\
\hline
\verb|CAUSE| & {\textit{Uvelichivsheesya potreblenie rafinirovannyh produktov pitaniya \textbf{yavlyaetsya prichinoj} mnozhestva takih zabolevanij.}} & Increased consumption of refined foods \textbf{is the cause} of many diseases. & {23}\\ 
\hline
\verb|ISA| & {\textit{Odnim iz samyh tochnyh i effektivnyh sposobov upravleniya zhestami \textbf{yavlyaetsya} upravlenie aktivnost'yu myshc.}} & One of the most accurate and effective ways to control gestures \textbf{is} to control muscle activity. & {13}\\
\hline
\verb|PART_OF| & {\textit{Process referirovaniya \textbf{sostoit iz} pyati osnovnyh shagov.}} & The referencing process \textbf{consists of} five main steps. & {5}\\ 
\hline
\verb|SYNONYMS| & {\textit{Stat'ya posvyashchena issledovaniyu vertikal'nogo poleta robota s mashushchim krylom, \textbf{takzhe nazyvaemogo} ornitopterom.}} & The article is devoted to the study of the vertical flight of a robot with a flap wing, \textbf{also called} an ornithopter. & {5}\\ 
\hline
\verb|TOOL| & {\textit{V stat'e predstavlen opyt razrabotki informacionnoj sistemy, \textbf{avtomatiziruyushchej} process raspredeleniya studentov po bazam praktik.}} & The article presents the experience of development of the information system, \textbf{which automates} the process of distribution of students by the bases of practice. & {29}\\ 
\hline
\verb|USAGE| & {\textit{V nastoyashchee vremya aktivno razvivaetsya napravlenie, svyazannoe s proektirovaniem nejronnyh setej \textbf{dlya ispol'zovaniya v} mobil'nyh ustrojstvah.}} & Currently, the development of neural networks \textbf{for use on} mobile devices is growing rapidly. & {36}\\\hline
\end{tabular}
\caption{Examples of lexical patterns}
\label{tab:lexicalpatterns}
\end{table*}



\subsection{Classification task with a CLS-vector}

The second approach we used for the relation extraction is similar to R-BERT, and is used by other authors \cite{hosseini-etal-2022-knowledge, aldahdooh_2021, pubmedKB}. We consider the task of relation extraction as a classification task (with 7 classes of relations: CAUSE, ISA, PART-OF, SYNONYMS, TOOL, USAGE, NO-RELATION). We take the vector of a special token CLS (it is considered as a vector of the input text) and the vector of two terms (connected by the relation). These three vectors are concatenated and the resulting vector is fed to the classifier \cite{Wu_2019}. We tried to use three different language models: mBERT, ruBERT \cite{ruBERT2019} and \href{https://huggingface.co/cointegrated/rubert-tiny2}{cointegrated/rubert-tiny2}. 

In addition, some features of the training are noteworthy. Firstly, to train the models, we used the corpus of Russian texts without dividing it into training and validation sets, and the most appropriate number of epochs was selected experimentally, because there are very few examples for some relations, and therefore the validation set would be unrepresentative to determine the quality of the model. Secondly, to reduce the imbalance between the number of examples in the classes, we added only 50\% of the randomly selected pairs of terms to the training set, excluding those with the distance between tokens more than 10.

Finally, we implemented an ensemble which includes both approaches: model and lexical patterns. All metrics are presented in Table~\ref{tab:metrics_models}. F1-score for all types of relations for combined approach are presented in Table~\ref{tab:f1_for_each_model}.  For comparison, the state-of-the-art result achieved on SciERC with the SpERT (using SciBERT) method is 50.84\% for relation extraction \cite{eberts-etal-2020-span}. Our results may also be related to insufficient data, as Russian is morphologically rich, which additionally complicates the work of the language model. Moreover, error analysis of relation extraction revealed that relations are often present implicitly between terms and one can recognize them if one only knows these particular terms. Quite many terms in IT texts are abstract (for example, \emph{"program implementation"}, \emph{"testing"}, etc.) and it can be difficult to define, whether there is any semantic relation between them or not. We plan to study this aspects in the future.

\begin{table}[hbt!]
\centering
\small
\begin{tabular}{llll}
\hline
\textbf{Model} & \textbf{Precision} & \textbf{Recall} & \textbf{F1}\\
\hline
\verb|mBERT| & {0.26} & {0.32} & {0.26}\\ 
\verb|ruBERT| & {0.26} & {0.34} & {0.27}\\
\verb|rubert-tiny2| & {0.22} & {0.23} & {0.22}\\ 
\hline
\verb|mBERT + p| & {0.26} & \textbf{0.41} & \textbf{0.29}\\ 
\verb|ruBERT + p| & \textbf{0.29} & {0.35} & {0.28}\\ 
\verb|rubert-tiny2 + p| & {0.29} & {0.24} & {0.24}\\
\hline
\end{tabular}
\caption{Metrics for different language models and combined approach; \emph{p} is for patterns}
\label{tab:metrics_models}
\end{table}

\begin{table}[hbt!]
\centering
\small
\begin{tabular}{llll}
\hline
\textbf{Relation type} & \textbf{mBERT} & \textbf{ruBERT} & \textbf{ruBERT-tiny2}\\
\hline
\verb|CAUSE| & {0.06} & {0.09} & {0.10}\\ 
\verb|ISA| & {0.30} & {0.28} & {0.14}\\
\verb|PART_OF| & {0.14} & {0.04} & {0.00}\\ 
\verb|SYNONYMS| & {0.32} & {0.33} & {0.38}\\ 
\verb|TOOL| & {0.04} & {0.07} & {0.00}\\ 
\verb|USAGE| & {0.27} & {0.22} & {0.11}\\
\verb|NO-RELATION| & {0.93} & {0.95} & {0.94}\\
\hline
\verb|macro-average| & \textbf{0.29} & {0.28} & {0.24}\\ 
\hline
\end{tabular}
\caption{F1-score for all types of relations for combined approach}
\label{tab:f1_for_each_model}
\end{table}

\section{Conclusion}
In this paper, we built a new dataset and study several methods for term recognition and relation extraction from computer science texts in Russian. We conducted several experiments with different pre-trained language models for both tasks. The results of our experiments show that language models pre-trained on the target language are not always show the best performance. Also adding some heuristic approaches may improve the overall quality for the particular task.

\bibliography{anthology,custom}
\bibliographystyle{acl_natbib}




\end{document}